\documentclass[lettersize,journal,twoside]{IEEEtran}

\usepackage{cite}
\usepackage{epsfig}
\usepackage{graphicx}
\usepackage{caption}
\usepackage{subcaption}
\usepackage{multirow}
\usepackage{booktabs, makecell, multirow}
\usepackage{amsmath}
\usepackage{amsfonts,amssymb}
\usepackage{url}
\usepackage{comment}
\usepackage{xcolor}
\usepackage{hyperref}
\DeclareMathOperator*{\argmin}{arg\,min}
\usepackage{pifont}
\newcommand{\cmark}{\ding{51}}
\newcommand{\xmark}{\ding{55}}

\usepackage{cite}
\usepackage{amsmath,amssymb,amsfonts}
\usepackage{algorithmic}
\usepackage{graphicx}
\usepackage{algorithm,algorithmic}
\usepackage{hyperref}
\usepackage{textcomp}

\begin{document}
\title{Time-Series U-Net with Recurrence for Noise-Robust Imaging Photoplethysmography}

\author{Vineet R. Shenoy\thanks{Vineet R. Shenoy is with the Johns Hopkins University Baltimore, MD, USA (e-mail: vshenoy4@jhu.edu). VS performed part of this work as an intern at MERL.}, Shaoju Wu\thanks{Shaoju Wu is with Boston Children’s Hospital and Harvard Medical School, Boston, MA USA. SW performed this work as an intern at MERL.}, Armand Comas\thanks{Armand Comas is with Google. AC performed this work as an intern at MERL.}, Tim K. Marks, Suhas Lohit, Hassan Mansour \thanks{Tim K. Marks, Suhas Lohit, and Hassan Mansour are with Mitsubishi Electric Research Laboratories (MERL), Cambridge, MA, USA (e-mail: \{tmarks, slohit, mansour\}@merl.com).}}

\markboth{}{Shenoy \MakeLowercase{et al.}: Time-Series U-Net with Recurrence for
Noise-Robust Imaging Photoplethysmography }

\IEEEpubidadjcol
\maketitle

\begin{abstract}
Remote estimation of vital signs enables health monitoring for situations in which contact-based devices are either not available, too intrusive, or too expensive. In this paper, we present a modular, interpretable pipeline for pulse signal estimation from video of the face that achieves state-of-the-art results on publicly available datasets. Our imaging photoplethysmography (iPPG) system consists of three modules: face and landmark detection, time-series extraction, and pulse signal/pulse rate estimation. Unlike many deep learning methods that make use of a single black-box model that maps directly from input video to output signal or heart rate, our modular approach enables each of the three parts of the pipeline to be interpreted individually. The pulse signal estimation module, which we call TURNIP (Time-Series U-Net with Recurrence for Noise-Robust Imaging Photoplethysmography), allows the system to faithfully reconstruct the underlying pulse signal waveform and uses it to measure heart rate and pulse rate variability metrics, even in the presence of motion. When parts of the face are occluded due to extreme head poses, our system explicitly detects such ``self-occluded" regions and maintains estimation robustness despite the missing information. Our algorithm provides reliable heart rate estimates without the need for specialized sensors or contact with the skin, outperforming previous iPPG methods on both color (RGB) and near-infrared (NIR) datasets.

\end{abstract}

\section{Introduction}

Health monitoring can improve the lives and increase the lifespans of all people, healthy or not. And with an increasing aging population globally, health monitoring can ease health difficulties for both the aging and their caregiers. From small commercial devices such as smart watches to large clinical machines such as CT scanners, medical devices keep physicians, patients, and everyday users aware of their health. These devices, however, are themselves sometimes a barrier to health monitoring---commercial devices tend to be expensive, while clinical devices are only accessible in medical facilities. Furthermore, medical equipment that requires physical contact with the body can be invasive and uncomfortable, limiting wider adoption of ubiquitous health monitoring technologies.

The COVID-19 pandemic has renewed interest in non-contact measurements of vital signs\cite{telemedicine, 10.1001/jamaneurol.2020.1452}, including pulse rate \cite{deephys, inverseCAN, poh, Poh_2010, metaphys, autosparseppg}, breathing rate \cite{exhalation, breathing, Cho_2017, thermistor}, blood pressure \cite{bp_estimation}, and pulse transit time~\cite{exhalation}. Remote healthcare has allowed patients to receive quality care even when offices are closed, leading to healthier and safer lives. Beyond healthcare, this type of monitoring could potentially be used in safety-critical applications such as driver-monitoring \cite{autosparseppg,sparseppg} and heavy equipment operation. Measuring quantities such as heart rate and pulse rate variability --- defined  as fluctuations in the interval between successive beats of a heart --- and doing so from facial video only can be used from hospital-based settings, to consumer electronics, and even safety-critical applications.

Imaging photoplethysmography (iPPG), also known as remote photoplethysmography (rPPG), is the process of determining the heart rate and/or pulse waveform from non-contact video of the skin (e.g., the face). The key to accurately estimating the pulse waveform is to first measure the variation in the image intensity of the skin that contains the underlying pulse signal, which is weak and noisy, then extract the pulse signal through denoising~\cite{pulse_webcam, poh} or signal processing-based estimation \cite{pbv, chrom, algorithmic_principles}. Many pre-deep-learning methods \cite{pulse_webcam, poh, pbv, chrom} first detect the face in each video frame, then average the RGB pixel intensities across the entire face region in each frame, to obtain a 3-channel (R, G, B) time series. These algorithms estimate the underlying pulse signal from the spatially averaged 3-channel signal either by de-mixing under model-free assumptions of blind source separation \cite{pulse_webcam, poh} or by projecting the RGB signals onto different color subspaces \cite{pbv, chrom}. Deep learning-based methods such as \cite{inverseCAN, deephys, tscan} input video frames directly and use attention modules to extract the pulse signal from the face region. 

The SparsePPG~\cite{sparseppg} and AutoSparsePPG~\cite{autosparseppg} methods first segment the face into regions and record the variation in the mean intensity value of each region over time. Unlike \cite{pulse_webcam, poh, pbv, chrom, algorithmic_principles}, which extract a 3-channel (R, G, B) time series from the video, these methods~\cite{sparseppg,autosparseppg} extract a multi-channel time series whose channels represent facial regions rather than colors. From this multi-channel time series, the heart rate is estimated by assuming that the pulse signal is sparse in the frequency domain and using sparsity-promoting algorithms to find the underlying heart-rate frequencies that are shared across facial regions. Our proposed method adopts a similar multi-region analysis, but rather than using sparsity-driven algorithms as in \cite{sparseppg, autosparseppg}, we develop a  network architecture that learns to extract the underlying pulse signal from the multi-region time series.

We adopt a modular framework for pulse signal estimation that achieves state-of-the-art results on publicly available datasets. We demonstrate the effectiveness of our algorithm in two imaging domains: the color (RGB) domain and the near-infrared (NIR) domain. Our model can recover the pulse rate  in the presence of substantial motion, and our system's detection and special handling of \textit{self-occluded} landmarks  (facial landmarks that are occluded by other parts of the face) makes our method more robust to extreme head poses such as profile views. In addition, we explore a pulse rate variability analysis, which reflects neurocardiac function and autonomic nervous system activity. Our model includes three modules: a face and landmark detection module, a time-series extraction module, and a pulse signal estimation module called TURNIP (\textbf{T}ime-Series \textbf{U}-Net with \textbf{R}ecurrence for \textbf{N}oise-Robust \textbf{I}maging \textbf{P}hotoplethysmography\footnote{An earlier version of this work appeared in \cite{turnip}, where the ``N" in TURNIP stood for Near-Infrared. In the current paper, we expand the algorithm to operate on RGB input, enable the handling of occlusions, and employ an improved face and landmark detector. We also evaluate our algorithm on a larger dataset and demonstrate improved performance relative to state-of-the-art deep-learning methods. We also add a pulse rate variability analysis to the paper.}). The system reconstructs the underlying pulse signal, which is then used for pulse rate and pulse rate variability estimation. The face detection and landmark detection module, as well as the TURNIP iPPG estimation module, contain deep-learning components, while the time-series extraction module explicitly records the temporal variation of the light intensity within each facial region. As a combined system, these three modules determine the pulse rate with state-of-the-art accuracy, even when head motion is present in the data.

To summarize, our contributions are as follows:

\begin{itemize}
    \item We design a modular, interpretable pipeline for pulse-rate estimation and pulse rate variability from face videos; this pipeline consists of a face and landmark detection module, a time-series extraction module, and a pulse signal estimation module.
    \item We propose TURNIP, a time-series U-Net with Gated Recurrent Units (GRUs) as pass-through connections, which reconstructs the underlying pulse signal.
    \item For color (RGB) videos, we demonstrate that the ratio of the red and green color channels \cite{chrom, RoverG-1, RoverG-2} in a spatio-temporal neural network better denoises input signals than raw color channels, leading to improved pulse signal estimation.
    \item We handle extreme head poses and poorly framed video by automatically identifying self-occluded or outside-of-frame landmarks, enabling our method to learn to handle bad information and be robust to extreme poses.
    \item We evaluate our algorithm on three publicly available datasets from the RGB and NIR domains, achieving state-of-the-art results on all datasets.
\end{itemize}

The rest of the paper is organized as follows: in Section~\ref{sec:RelatedWork}, we discuss related work. This is followed by our pulse signal and pulse rate estimation technique in Section~\ref{sec:Method}. The implementation details and experimental results are described in Section~\ref{sec:Experiments}, and we conclude in Section~\ref{sec:conclusion}.


\section{Related Work}\label{sec:RelatedWork}

\subsection{Pulse Rate Estimation}
Pulse Signal estimation can be separated into signal processing-based methods \cite{pulse_webcam, poh, pbv, chrom, sparseppg, autosparseppg, tulyakov} and deep neural network methods~\cite{inverseCAN, deephys, contrastphys, metaphys, rppg_instr, liu2018remote, lee2020meta, instr2, du2023dual, rppg-mae, song2021pulsegan, synrhythm}. We discuss each individually below.

\subsubsection{Signal Processing-Based Methods}

Signal processing-based methods include blind source separation (BSS) methods such as \cite{pulse_webcam, poh} and model-based methods such as \cite{pbv, chrom, sparseppg}.
Blind Source Separation techniques \cite{pulse_webcam, poh} consider the measured signal to contain both the underlying pulse signal and noise. To separate the signal from the noise, they use Principal Component Analysis (PCA) or Independent Component Analysis (ICA), depending on whether they desire the projected data to lie in the coordinate systems of maximum variance or independence. Unlike these BSS methods, which do not consider skin-reflection models such as~\cite{algorithmic_principles}, CHROM~\cite{chrom} explicitly considers a subject's skin color as well as the light source upon the skin. It does so by eliminating the specular reflection component and white-balancing the underlying pulse signal using a standardized skin-tone vector. PBV~\cite{pbv} restricts all color variations to the pulsatile direction, assumes that the pulsatile signal is uncorrelated with other signal sources, and solves for a projection vector to obtain the pulsatile signal. 

The methods SparsePPG~\cite{sparseppg} and AutoSparsePPG~\cite{autosparseppg} are built upon sparse recovery algorithms. Recognizing that the pulse-rate signal is quasi-periodic in time and sparse in the frequency domain, these methods seek to recover the peak frequency coefficients of the pulse signal. After segmenting the face into five regions and extracting a time-series from pixel intensities across video frames, these methods solve an optimization problem in which they seek to extract the underlying sparse set of frequency coefficients that describe the periodicity of the pulse signal. They assume that the set of active frequencies that correspond to the pulse signal should be the same across the five facial regions, and they use techniques in joint sparsity to solve for the target signal.

\begin{figure*}[ht]
    \centering
    \includegraphics[width=0.98\textwidth]{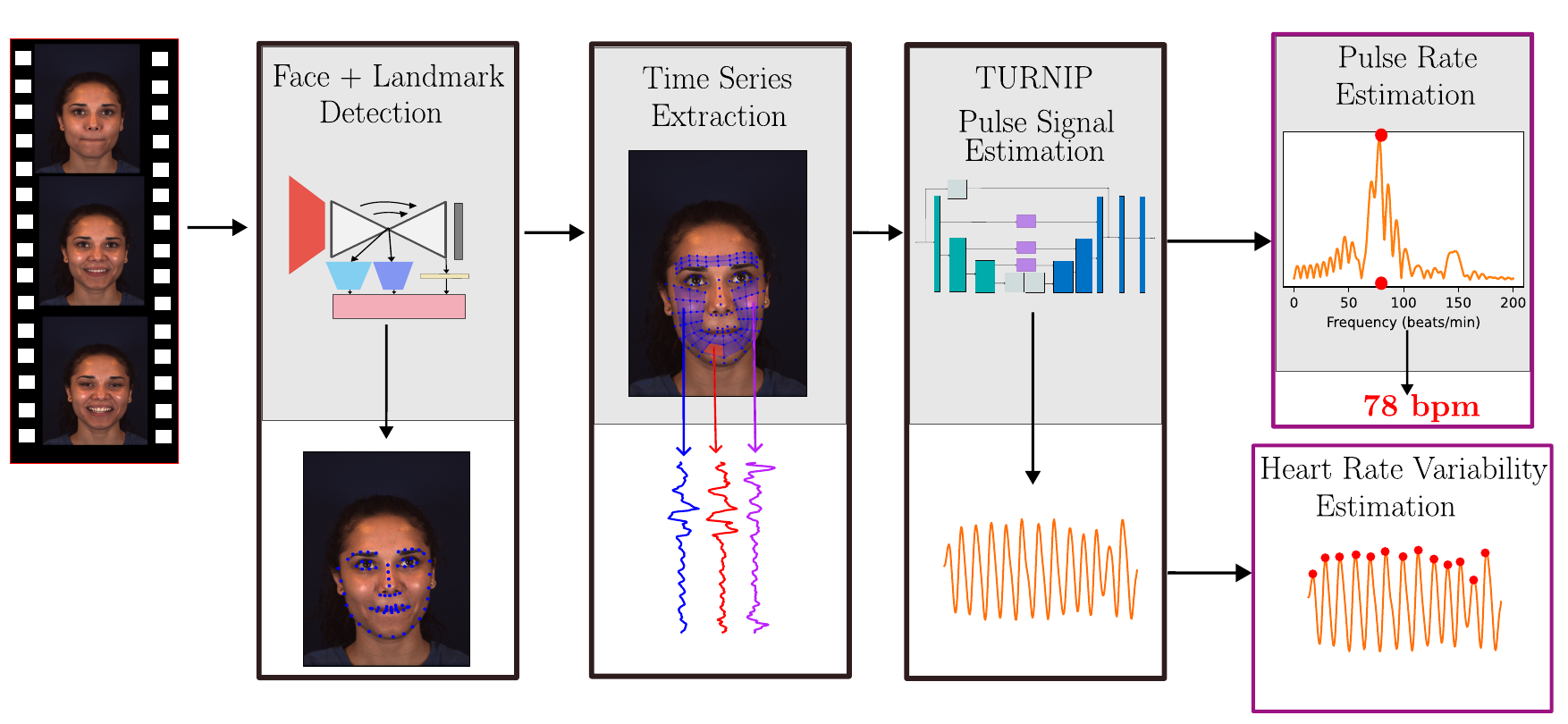}
    \caption{Our system for pulse signal estimation from video is composed of three modules, outlined in black: face and landmark detection, time series extraction, and pulse signal estimation. The pulse rate and pulse rate variability can then be estimated from the denoised pulse signal that is output from TURNIP.}
    \label{fig:heart-rate-pipeline}
\end{figure*}

\subsubsection{Deep-Learning Methods}

Purely signal processing-based methods have recently been surpassed by deep learning-based methods. PhysNet\cite{physnet} was among the first end-to-end methods to use spatio-temporal neural networks to reconstruct a pulse waveform directly from raw RGB video. Some methods, such as~\cite{deephys}, obviate the need for explicitly-defined signal extraction techniques such as those described in the previous paragraph. Using the skin reflection model defined in~\cite{algorithmic_principles}, \cite{deephys} inputs the difference between consecutive frames, rather than the frames themselves, to eliminate the stationary skin reflection color. The algorithm then uses the MSE loss between this difference signal and the corresponding difference signal of the ground-truth waveform. To account for motion, an attention mechanism is developed using soft-attention masks from $1 \times 1$ convolutions that are multiplied with the motion model feature map; the result only highlights the skin region for signal extraction. Follow-up work \cite{inverseCAN} improves on this attention mechanism by claiming that changes in background and ``distraction" regions (such as the hair) can improve the quality of the attention mechanism to focus more on skin pixels. The work TS-CAN~\cite{tscan} added temporal shift modules~\cite{tsm} to \cite{deephys} for better temporal processing, and MetaPhys \cite{metaphys} used this network in a meta-learning paradigm for few-shot adaptation across datasets in both supervised and unsupervised training procedures. A benefit of \cite{deephys, inverseCAN, tscan, metaphys, contrastphys} is that these methods are end-to-end: RGB video frames are input, and the pulse signal is output. As a result, however, these methods become hard to interpret. 

\subsection{Pulse Rate Variability estimation}
In addition to measuring the heart rate, we explore our algorithm's capability to measure pulse rate variability (PRV) metrics. Pulse Rate Variability, defined by the time interval between adjacent heart beats and the fluctuations between them, measures the heart-brain dynamic interactions and is closely associated with autonomic nervous system activity~\cite{shaffer2017overview}. Fluctuation in the inter-beat-interval (IBI) are due to the dynamic relationship between the sympathetic nervous system and the parasympathetic nervous system, as well as respiratory sinus arrhythmia and changes in tone of the vasculature~\cite{shaffer2017overview}. Measuring these changes through pulse rate variability can give insight into functioning of the heart, intestines, blood pressure, and more.

Measuring pulse rate variability, however, is highly dependent on the type of signal recorded (e.g. ECG vs PPG), the age of the subjects, and specifically, the length of the time recording. While 24-hour signal recordings are considered to be the ``gold-standard" for measurement, shorter length signal may or may not correspond with the 24 hour measurements. To this end \cite{salahuddin2007ultra, aeschbacher2017heart, baek2015reliability, munoz2015validity, shaffer2016promise} have established correlations between shorter duration measurements and the gold-standard 24-hours recording. We follow the work of~\cite{baek2015reliability} detailing PRV for short term measurements using PPG waves, and report the power of the High Frequency (\textbf{HF}) components of the interbeat interval signal (in ms$^2$) and the root mean square of successive differences between normal heartbeats (\textbf{RMSSD}) measured in milliseconds (ms). For a full description of pulse rate variability, we refer the reader to \cite{shaffer2017overview}.


\section{Method}\label{sec:Method}
Our heart rate estimation framework is composed of three modules, outlined in black in Figure~\ref{fig:heart-rate-pipeline}: a face and landmark detector, a time-series extractor , and a pulse signal estimator. Each component is described in the following subsections.

\subsection{Face and Landmark Detection}\label{subsec:face_landmark_detection}

Given the raw, unprocessed video, we detect the face box in each frame using~\cite{Faceboxes}, and the cropped faces are input to the LUVLi landmark detection algorithm~\cite{luvli}. The key advantage of LUVLi for this application is that in addition to estimating landmark locations, it indicates which (if any) landmarks are occluded due to head pose, labeling them as {\em self-occluded}. We consider a landmark to be {\em invisible} if it is self-occluded as shown in Figure~\ref{fig:bad_landmarks.pdf}. Given that pulsatile signals in rPPG are weak, labeling landmarks as invisible is important to alert downstream modules of any noise that may corrupt the underlying PPG signal. We use these labels in generating the time series that are input to TURNIP, as described below. 

\subsection{Time Series Extraction}\label{subsec:timeseries_extraction}

\subsubsection{Generating Additional Landmarks}\label{subsubsection:lanmarks}
\begin{figure*}[t]
    \centering
    \includegraphics[width=0.75\textwidth]{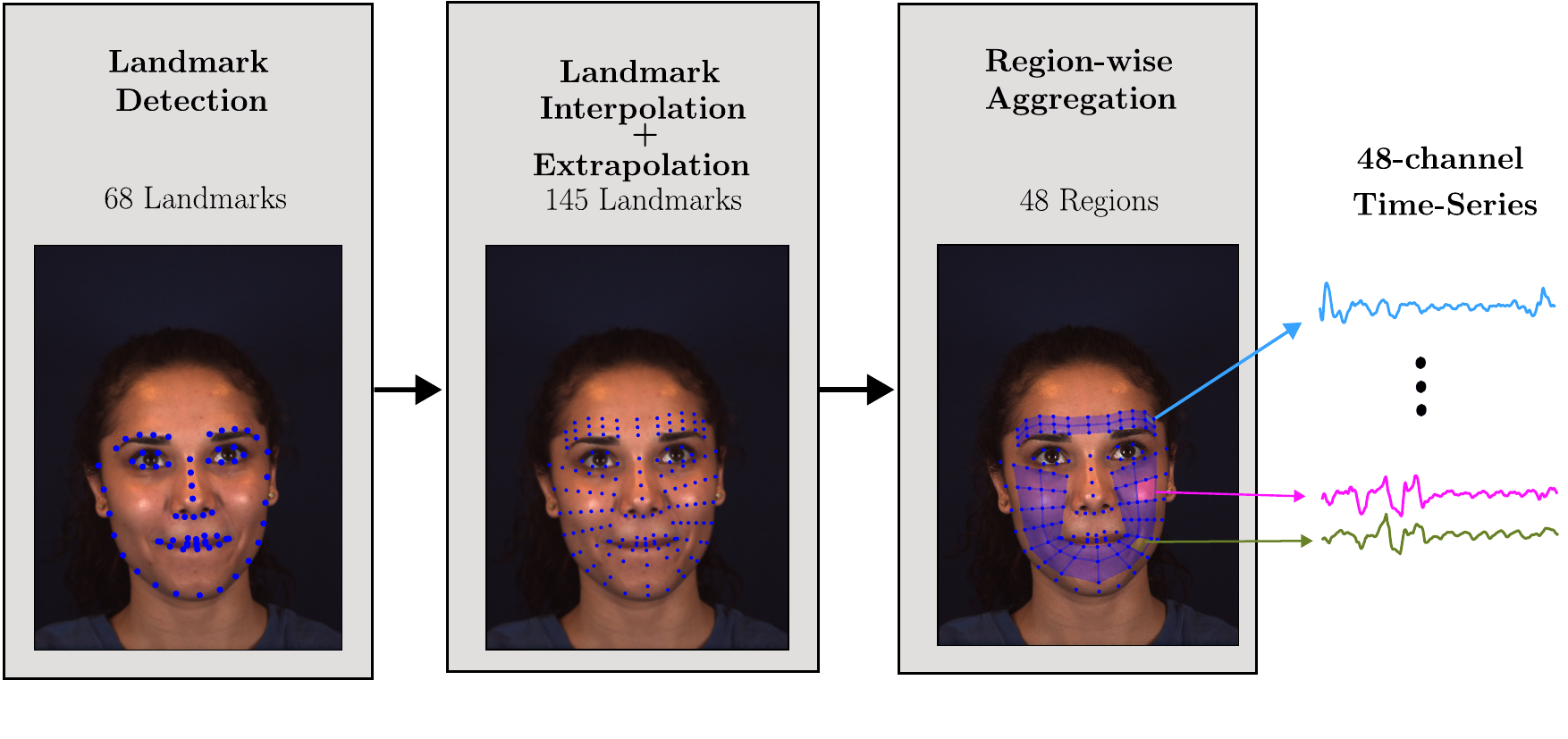}
    \caption{Generating of landmark and feature regions. We first start by detecting 68 landmarks from the LUVLi \cite{luvli} landmark detector. We then interpolate these landmarks across the cheeks and chin and extrapolate them up the forehead, to generate 145 landmarks. We use these landmarks to define 48 regions. Finally, we aggregate the pixel intensities in each region using spatial averaging to obtain a 48-channel time-series.}
    \label{fig:landmark_progression.pdf}
\end{figure*}

\begin{figure}[t]
    \centering
    \includegraphics[width=0.49\textwidth]{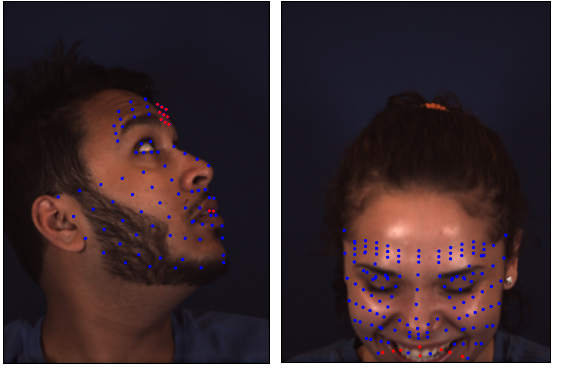}
    \caption{These example frames show the augmented set of 145 landmarks, with {\em invisible} landmarks shown in red.  Landmarks are also labeled {\em invisible} if they are self-occluded. Finally, landmarks are labeled invisible if their locations were determined by interpolating/extrapolating using an invisible landmark. Previous algorithms such as \cite{sparseppg,autosparseppg} do not detect when landmarks are invisible, which means that such landmarks can cause previous methods to have incorrect results in frames that have extreme head rotations or translations. Because we explicitly detect invisible landmarks and label them as such, our algorithm can learn to be more robust to extreme poses.}
    \label{fig:bad_landmarks.pdf}
\end{figure}

Given the landmark points in each frame and their visibility labels, we will extract the temporal pulsatile signal, potentially noisy, at different regions in the face.
To improve signal reconstruction, we augment the number of landmarks by interpolating/extrapolating new landmarks on the cheeks, chin, and forehead as illustrated in Figure~\ref{fig:landmark_progression.pdf}. The new landmarks on the cheeks and chin are obtained by linearly interpolating between two existing landmarks. For example, we interpolate between the lower lip and the jawline to get chin landmarks. To extrapolate landmarks on the forehead, we first calculate a direction vector $\mathbf{v}_{\text{eyebrow}}$ along the right and left eyebrows by fitting a line to all visible eyebrow landmarks, then compute the perpendicular vector $\mathbf{v}_{\text{forehead}}$. We then extrapolate two rows of landmarks in the direction of $\mathbf{v}_{\text{forehead}}$, using one-fifth the distance between the inside eye corner and the corresponding mouth corner as the offset for each row. If one of the the 68 landmarks is invisible (as defined in Section \ref{subsec:face_landmark_detection}), then we propagate its {\em invisible} label to every one of the augmented landmarks whose locations were determined using the invisible landmark. This is illustrated by the examples in Figure~\ref{fig:bad_landmarks.pdf}.  Landmark augmentation from 68 to 145 landmarks helps to capture more regions of the face, including the critical forehead regions; this is particularly helpful in cases in which the pulsatile signal is weak or noisy in other regions.

\subsubsection{Handling Pixel Intensities}\label{subsubsection:pixel_intensities} We incorporate three design features into the time series extraction module that contribute to our state-of-the-art results: \textbf{1)} We use these 145 landmark locations in each frame to define 48 facial regions, and within each region we average the pixel intensity values (e.g., Red-over-Green ratio) across all pixels in the region. This step reduces noise. \textbf{2)} If a region is defined using an {\em invisible} landmark, we assign that region a large out-of-range value instead of averaging its intensity; this step recognizes that some signals will be inherently incorrect and noisy, and should be ignored or used for noise-robustness. \textbf{3)} For the RGB datasets (MMSE-HR and PURE), instead of extracting pixel intensities from a single color channel (e.g., the green channel of an RGB image), we extract a channel defined by the red-divided-by-green (R/G) intensity ratio; this improves signal strength and reduces the effects of motion noise, which we show empirically in Section \ref{sec:Experiments}.

\subsubsection{Filtering and Normalizing the Signals}\label{subsubsection:normalize_filter}Finally, we temporally filter each of the 48 spatial regions using a fifth-order Butterworth filter with cutoff frequencies at 0.7 Hz and 4 Hz as in \cite{inverseCAN}, corresponding to 42 beats per minute (bpm) and 240 bpm, respectively. Additionally, we normalize the signals to the range $[-1, 1]$, and perform AC/DC normalization by subtracting the signal's mean from the signal, then dividing by the same mean:
\begin{equation}
\mathbf{\hat{y}}_i = \frac{(\mathbf{y}_i - \mu_i)}{\mu_i},
\end{equation}
where $\mathbf{y}_i$ is the signal from region $i$, and $\mu_i$ is the temporal mean intensity in region $i$. The resulting 48-channel time series is input to the TURNIP denoiser network, which estimates the pulse signal.

\subsection{TURNIP Pulse Signal Estimation} \label{subsec:denoising}

\begin{figure*}[ht]
    \centering
    \includegraphics[width=0.99\textwidth]{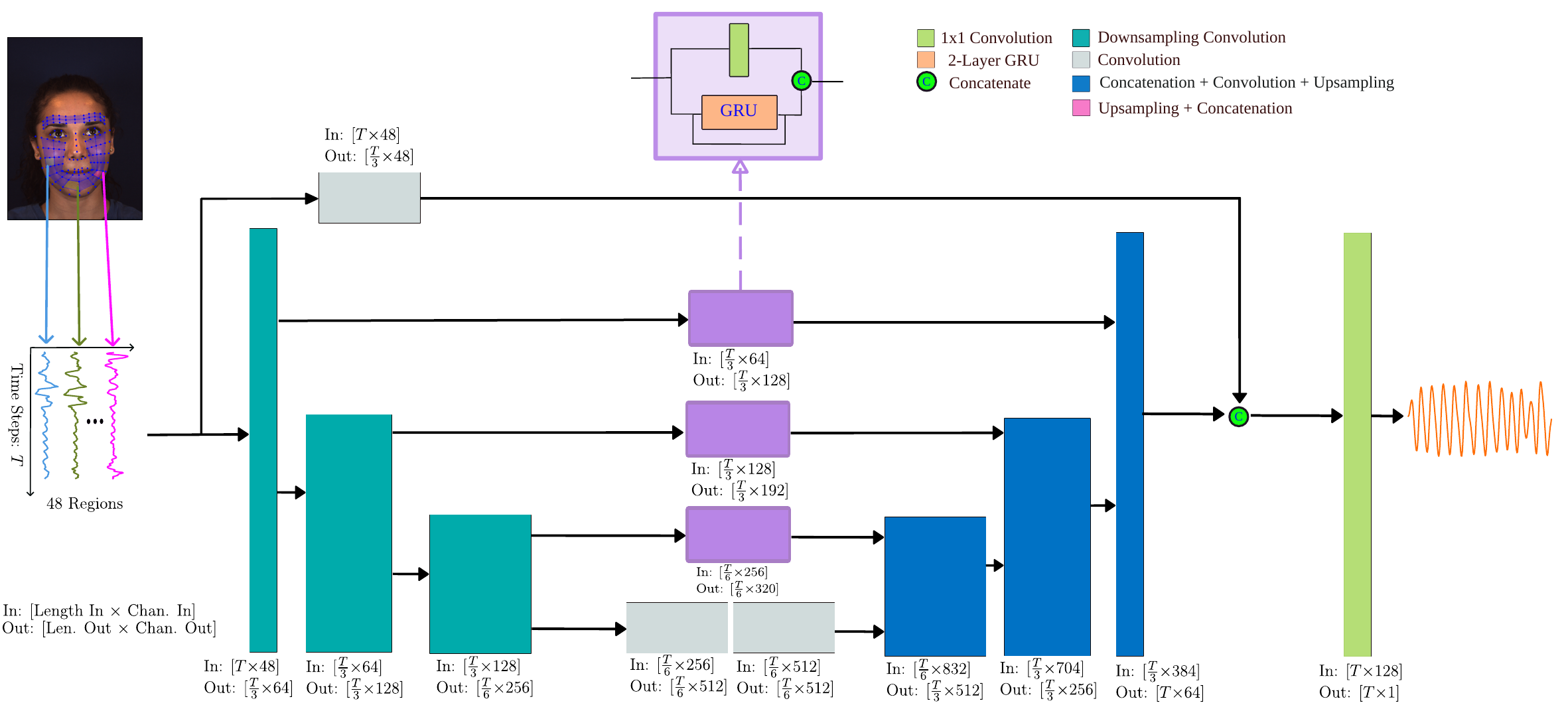}
    \caption{The TURNIP Pulse Signal Estimation module. The signals from the 48 individual regions are extracted at input to the network as a $T\times48$ matrix. The spatio-temporal network denoises the signal to the statistics of the data, and outputs a clean signal. See Figure \ref{fig:three_iter} for example input and output of TURNIP. }
    \label{fig:turnip.pdf}
\end{figure*}

The extracted time series is windowed into $T \times 48$ chunks for input into the TURNIP module, where $T$ is the length of the time window. Segmenting the skin pixels into regions and spatially averaging them reduces noise; however, noise is still present due to facial deformations resulting from expressions, motion-related noise, and lighting variations, among other sources of noise. The network must learn to extract the pulse signal based on the statistics of the ground-truth data.

We design our architecture as a U-Net style neural network~\cite{unet}. However, we modify the skip connections of the U-net to incorporate temporal recurrence, in the form of gated recurrent units (GRUs). The  architecture of our pulse signal estimation module, which we call Time-Series U-net with Recurrence for Noise-Robust Imaging Photoplethysmography (TURNIP), is shown in Figure \ref{fig:turnip.pdf}.

The $T \times 48$ signal is passed to the neural network as 48 channels. It goes through three stages of convolution (with kernel size 7) and downsampling, first by a factor of three and then by a factor of two. After reaching the lowest resolution, the signal is convolved and upsampled back to the original spatial and channel dimension over multiple stages. At every resolution of the U-net, we connect the encoding and decoding sub-networks by a skip connection module (each indicated by a purple rectangle in Figure~\ref{fig:turnip.pdf}). In parallel with each $1 \times 1$ convolutional skip connection (which is present in a typical U-net), we introduce a novel recurrent skip connection that uses gated recurrent units (GRUs) to provide temporally recurrent features. The output (hidden states) of this GRU layer is concatenated with the output of the standard ($1 \times 1$) skip connection layer before being concatenated with the input to the corresponding convolution+upsampling layer. At each time scale, the convolutional layers of the U-net process all of the samples from the time window in parallel. In contrast, the new recurrent GRU layers process the temporal samples sequentially. This recurrence effectively extends the temporal receptive field at each layer of the U-net. After the last upsampling layer, a 1x1 convolutional layer is appended to collapse the channel dimension to a single channel for appropriate calculation of a loss.

Training the TURNIP pulse signal estimation module requires an appropriate selection of loss function. We chose to minimize one minus the Pearson Correlation Coefficient, a measure of covariance between two variables. Consider two vectors $\mathbf{\bar{z}}$ and $\mathbf{z}_{\text{gt}}$ representing the predicted waveform and the ground-truth waveform of length $T$, respectively. We define a function $F(\mathbf{\bar{z}}, \mathbf{z}_{\text{gt}})$ such that 

\begin{equation}
    F(\mathbf{\bar{z}}, \mathbf{z}_{\text{gt}}) = 1 - \frac{T \cdot \mathbf{\bar{z}}^\intercal  \mathbf{z}_{\text{gt}} - \mu_{\bar{z}}\mu_{z_{\text{gt}}}}{\sqrt{(T \cdot \mathbf{\bar{z}}^\intercal \mathbf{\bar{z}} - \mu_{\bar{z}}^2) (T \cdot \mathbf{{z}}_{\text{gt}}^\intercal \mathbf{z}_{\text{gt}} - \mu_{z_{\text{gt}}}^2)}}
\end{equation}

We then seek to minimize this function with respect to the parameters of the neural network

\begin{equation}
    \theta^* = \argmin_{\theta} F(\mathbf{\bar{z}}, \mathbf{z}_{\text{gt}})
\end{equation}

During training, we also design a novel data augmentation scheme that aims to capture lower and higher frequencies of the target pulse rate range, for which sufficient training data may not be available. In our ``SpeedUp" augmentation, we crop the input window of length $T$ by a random percentage between 20\% and 40\%, and linearly interpolate the samples back to the original window size $T$. For the ``SlowDown" augmentation, we randomly chose a length that is 20\% to 40\% larger than our target time windows (e.g. $1.2 \times T$), extract the signal from the video, and linearly interpolate the signal to the target length $T$. In these ways our ``SpeedUp" and ``SlowDown" augmentation have extrapolated the statistics of the data to higher and lower frequencies respectively.

In section \ref{sec:Experiments}, we show that our formulation produces state-of-the-art results on major datasets, and we conduct ablation studies to demonstrate the effectiveness of our design choices.


    \begin{table}[]
    \centering
    \caption{Testing on three datasets in both the RGB and NIR domains}
    \label{tab:datasets}
    \resizebox{0.48\textwidth}{!}{\begin{tabular}{|c|ccc|}
    \hline
   & \textbf{MMSE-HR}~\cite{mmsehr} & \textbf{MR-NIRP Car}~\cite{autosparseppg} & \textbf{PURE}~\cite{Stricker2014NoncontactVP} \\
    \hline
    Domain & RGB & NIR & RGB \\
    Resolution & 1040$\times$1392 & 640$\times$640 & 640$\times$480\\
    Frame Rate (fps) & 25 & 30 & 30 \\
    Codec & h264 & h264 & h264 \\
    No. Videos & 102 & 19 & 60 \\
    No. subject & 40 & 18 & 10 \\
    Male/Female Subjects & 17/23 & 16/2 & 8/2 \\
    \hline
    GT Signal Type & BP Wave & Pulse Ox & Pulse Ox \\
    GT Sampling Rate (Hz) & 1000 &  60 & 60 \\
    \hline
    \end{tabular}}
    \end{table}

\section{Datasets}\label{sec:Datasets}
We test our algorithm on three video datasets in both the RGB and Near-Infrared (NIR) modalities, and describe the datasets in Table~\ref{tab:datasets} and below, as well as the the challenges associated with each dataset. We train and test on each dataset independently, except for the cross-dataset evaluation described in Section~\ref{sec:cross_datset}. For each dataset, we downsample the ground-truth signal to the frame rate of the video. We then apply the AC-DC normalization, $L_2$ normalization, and bandpass filtering described in Section~\ref{subsubsection:normalize_filter}.

\textbf{MMSE-HR}\cite{mmsehr}: In the MMSE-HR dataset, various emotions are elicited from subjects while ground-truth blood-pressure waveforms (synchronized with the video) are captured using a finger sensor that was calibrated with a blood pressure cuff. This dataset contains considerable head motion and occlusions, to which our algorithms are robust. During training, we used 10-second windows ($T=250$ samples), and shift the time window by 60 samples. Since we are performing leave-one-subject-out cross validation, this results in an average of 8026 windows for training and 9.7 windows for testing. During test time, we evaluate our model on 10-second samples with no overlap. We concatenate three 10-second windows and perform evaluation on 30-second windows to match the 30-second window length of~\cite{inverseCAN}.

\textbf{MERL-Rice Near-Infrared Pulse (MR-NIRP) Car} \cite{autosparseppg}: Recorded using an NIR camera with a $940 \pm 5$ nm bandpass filter, the MR-NIRP Car dataset includes facial videos (duration 2--5 min.) with synchronized ground-truth PPG waveforms collected using a fingertip pulse oximeter. The dataset is split into a ``Driving" subset and ``Garage" subset. During ``Driving", data were captured while driving through a city, resulting in illumination changes and head motion. There were 14 daytime videos and 4 videos that were captured at night. The ``Garage" subset was recorded while the car was parked inside a garage, with less head motion and much less variation in illumination. As in ~\cite{autosparseppg}, we evaluate only on the ``minimal head motion condition" for all scenarios -- we note, however, that even this scenario contains significant head motion in the Driving subset.  An advantage of the NIR modality is that it minimizes illumination variations~\cite{autosparseppg}. However, NIR frequencies introduce new challenges for iPPG, including weaker blood-flow-related intensity changes in the NIR portion of the spectrum and low signal-to-noise ratio (SNR) due to reduced sensitivity of camera sensors. The results demonstrate that despite these challenges, our method is able to accurately estimate subjects' heart rates.

\textbf{PURE}\cite{Stricker2014NoncontactVP}: In the PURE dataset, subjects perform various head motion tasks while synchronized video and pulse waveform data (from a fingertip pulse oximeter) are captured. There are six head-motion tasks: \textit{Steady, Talking, Slow Translation, Fast Translation, Small Rotation,} and \textit{Medium Rotation}. We split the dataset into train/validation/test/splits as in~\cite{Spetlik2018VisualHR}, resulting in 7613 training windows and 136 test windows.

\section{Experiments and Results}\label{sec:Experiments}
\subsection{Evaluation Protocol}
\subsubsection{Heart Rate Estimation}
To compute the heart rate estimate, we first multiply the time-series by a hanning window. We then
take the $L$-point FFT, where $L = 100 \times$signal length. We square the magnitude of the coefficients to
get the power, and take the bin with highest power (among the positive frequencies) as our estimate of the heart rate.

We follow the evaluation protocols as described in \cite{inverseCAN} and report the Mean Absolute Error (MAE) and Root Mean Squared Error (RMSE) between the predicted heart rate and the ground-truth heart rate. The MAE is defined as
\begin{equation}
    \frac{1}{N} \Sigma_{i=1}^N |R_i - \hat{R}_i|
\end{equation} and the RMSE is defined as 

\begin{equation}
    \sqrt{\frac{1}{N} \Sigma_{i=1}^N (R_i - \hat{R}_i)^2}
\end{equation}
 where $R_i$ is the ground-truth heart rate in time window $i$ , $\hat{R}_i$ is the predicted heart rate in  time window $i$, and $N$ is the total number of time windows. In addition, we report the PTE6 (\textbf{P}ercent of \textbf{T}ime that \textbf{E}rror~$<$~\textbf{6}~bpm), which is defined as 

\begin{equation}
    \text{PTE}6 = \frac{100}{N} \Sigma_{i=1}^N P_i, \text{ where } P_i =  
\begin{cases}
    1,& \text{if } |R_i - \hat{R}_i| < 6\\
    0,              & \text{otherwise}
\end{cases}
\end{equation}

 This quantity describes the percentage of heart rate estimates that are within 6 beats per minute (bpm) of the ground-truth heart rate. We chose this metric as it roughly encodes the notion of what percent of the time the estimated heart rate is correct. 

For a fair comparison against previous methods, we evaluate on 30-second time windows for the MMSE-HR dataset in accordance with evaluation protocols from~\cite{inverseCAN}, and on 10-second time windows for the MR-NIRP Car dataset to conform to our evaluation protocols in~\cite{turnip}. We evaluate on 30-second windows on the PURE dataset to conform with previous literature.

\subsubsection{Pulse Rate Variability Metrics}\label{sec:hrv_analysis} We report the metrics as described in~\cite{shaffer2017overview, baek2015reliability} which enumerate the ultra-short duration metrics that correlate well with the 24-hours recordings. Given that signals in the PURE dataset are approximately 1 min long, we report the power of the High Frequency (HF) components of the interbeat interval in milliseconds squared (ms$^2$) and the root mean square of successive differences between normal heart beats (RMSSD) in milliseconds (ms)~\cite{shaffer2017overview}. We report these numbers using the HeartPy~\cite{van2018heart, van2019analysing} python library, which standardizes the computation of the above metrics. We do not report the low frequency components of the interbeat interval signal as this metrics needs at least 2 minutes of recording,  nor the low-frequency/high-frequency ratio as this metric is most accurately reported on signals of 24-hr duration.

\subsection{Implementation Details}

For face detection, as described in Section \ref{subsec:face_landmark_detection}, we use the off-the-shelf detector as trained by \cite{Faceboxes}. The time series extraction was described in detail in Section \ref{subsec:timeseries_extraction}. Below, we describe the training procedure for the landmark detection introduced in Section \ref{subsec:face_landmark_detection} for the MMSE-HR results and the TURNIP iPPG estimator described in Section \ref{subsec:denoising}.

\textbf{LUVLi Landmark Localization}: In addition to accurate locations of facial landmarks, the LUVLi landmark detector~\cite{luvli} outputs a visibility for each landmark that indicates whether the landmark is ``self-occluded" (signifying that the landmark is occluded by another part of the face, e.g., in the case of a profile face). If LUVLi determines that a landmark is invisible (self-occluded or outside of the image boundaries), then for every face region that is defined using that landmark's location, our time series extraction module sets the region's intensity value to -10, which is a large negative value outside of the normalized range of the signal. This enables TURNIP to learn to ignore those regions when estimating the pulse signal. We show that this novel use of landmark visibility serves as a pseudo-attention mechanism that improves pulse signal estimation.

\textbf{TURNIP iPPG Estimation}: We use the architecture shown in Figure \ref{fig:turnip.pdf}. In consists of a three stage U-Net with 48 input channels; the time-resolution decreases while the channel dimension doubles at each stage until we reach 512 channels; we then decode this input by increasing the time-resolution and decreasing the channel resolution, adding GRU output along the way. The hidden state of the GRU is re-initialized for each time window of length $T$ that is fed into the network. For the MMSE-HR dataset, we use the Adam optimizer with an initial learning rate of $1.5e-3$ and weight decay of $1e-4$. We found the same hyperparameters worked well for PURE. On the MR-NIRP dataset, we use a learning rate of $1.5e-4$ reduced at each epoch by a factor of 0.05. The learning rate is decayed by a factor of $0.99$ at each epoch, and we train for 8 epochs. We use 10-second time window, and shift the window by 60 samples to generate our training set. Given the limited data, we use leave-one-subject-out cross validation for MMSE-HR and MR-NIRP Car datasets, and the train/val/test splits of the PURE dataset.

During test time, to replicate the 30-second-window evaluation process as described by~\cite{inverseCAN} on the MMSE-HR dataset, we concatenate three 10-second output windows of TURNIP to form a 30-second evaluation window. On the MR-NIRP Car dataset, we follow the evaluation protocol we described in~\cite{turnip}, which uses the same metrics as in~\cite{inverseCAN} but evaluates on 10-second windows. To be consistent with~\cite{turnip} on the MR-NIRP Car dataset, we use the OpenFace landmark detector~\cite{zadeh2017convolutional, baltruvsaitis2016openface}, smooth the resulting landmark locations using a 10-frame moving average, and do not use landmark visibility information (which is not available in OpenFace). In the next subsection, we show that our model achieves state-of-the-art performance on both datasets.

\begin{table}[]
    \centering
    \caption{Results on the MMSE-HR dataset using 30-second windows (TURNIP results show mean and standard deviation across four random network initializations). The results for ICA~\cite{Poh_2010}, CHROM~\cite{chrom}, and POS~\cite{algorithmic_principles} are copied from~\cite{inverseCAN}. AutoSparsePPG uses our signal extraction techniques, while~\cite{deephys, inverseCAN} have no analogous signal extraction technique.}
    \label{tab:mmsehr_results}
    \begin{tabular}{|c|ccc|}
    \hline
    \multirow{2}{*}{\textbf{Method}} & \textbf{MAE} & \textbf{RMSE} & \textbf{PTE6} \\
     & (bpm) $\downarrow$ & (bpm) $\downarrow$ & (\%) $\uparrow$ \\
    \hline
    
    ICA \cite{Poh_2010} & 5.44 & 12.00 & - \\
    
    CHROM \cite{chrom} & 3.74 & 8.11 & -\\
    POS \cite{algorithmic_principles} & 3.90 & 9.61 & - \\
    AutoSparsePPG \cite{autosparseppg} & 4.55 & 14.42 & 88.10 \\
    CAN\cite{deephys} & 4.06 & 9.51 & - \\
    InverseCAN \cite{inverseCAN} & 2.27 & 4.90 & -  \\
    Federated~\cite{liu2022federated} & 2.99 & - & 0.79 \\
    Physformer~\cite{yu2022physformer} & 2.84 & 5.36 & - \\
    EfficientPhys-C~\cite{liu2023efficientphys} & 2.91 & 5.43 & - \\
    ND-DeeprPPG~\cite{ndrppg} & 1.84 & 4.83 & - \\
    \hline
    
    \textbf{TURNIP}  & \textbf{1.17} & \textbf{3.46} & \textbf{93.21}\\
    \hline
    \end{tabular}
\end{table}

\begin{table}[]
    \centering
    \caption{Results on the PURE dataset. The results listed are based on implementations of \cite{dualgan}}
    \label{tab:pure_results}
    \begin{tabular}{|c|ccc|}
    \hline
    \multirow{2}{*}{\textbf{Method}} & \textbf{MAE} & \textbf{RMSE} & \textbf{PTE6} \\
     & (bpm) $\downarrow$ & (bpm) $\downarrow$ & (\%) $\uparrow$ \\
    \hline

    CHROM \cite{chrom} & 2.07 & 9.92 & -\\
    POS \cite{algorithmic_principles} & 5.44 & 12.00 & - \\
    HR-CNN~\cite{Spetlik2018VisualHR} & 1.84 & 2.37 &  -\\
    CVD~\cite{niu2020video} & 1.29 & 2.01 & - \\
    Gideon~\cite{Gideon_2021_ICCV} & 2.1 & 2.6 & - \\
    DualGAN~\cite{dualgan} & 0.82 & 1.31 & - \\
    Yue et. al~\cite{yue2023facial} & 1.23 & 2.01 & - \\
    ContrastPhys~\cite{contrastphy} & 0.48 & 0.98 & - \\
    
    \hline
    \textbf{TURNIP}  & \textbf{0.36} & \textbf{0.67} & \textbf{100}\\
    \hline
    \end{tabular}
\end{table}

\subsection{Results}

\subsubsection{Heart Rate Estimation Analysis}We show our results on the MMSE-HR dataset in Table~\ref{tab:mmsehr_results}.  On this challenging dataset, we reduce the Mean Absolute Error from the previous state of the art \cite{inverseCAN} from 1.84 to 1.17 bpm, and we reduce the RMSE error from 4.83 to 3.46 bpm. Furthermore, compared to the previous deep-learning-based methods~\cite{inverseCAN, deephys}, our modular system is more interpretable, as it does not involve the black-box attention mechanisms and signal reconstruction of these end-to-end approaches. Our pipeline has interpretable inputs and outputs that show exactly how a signal is extracted and how the underlying pulse waveform is estimated, which is a significant advantage over purely end-to-end methods. We see similar improvements on the PURE dataset, as shown in Table~\ref{tab:pure_results}. We exceed performance on both signal processing-based methods as well as deep-learning methods. In addition, we perform a cross-dataset performance study, which we defer to the ablation studies.

\begin{table}[]
    \centering
    \caption{Comparison of on the MMSE-HR dataset using 10-second vs. 30-second windows}
    \label{tab:10sec_mmsehr_results}
    \begin{tabular}{|c|ccc|}
    \hline
    \multirow{2}{*}{\textbf{Method}} & \textbf{MAE} & \textbf{RMSE} & \textbf{PTE6} \\
     & (bpm) $\downarrow$ & (bpm) $\downarrow$ & (\%) $\uparrow$ \\ 
    \hline
    TURNIP (30-sec) & 1.17$\pm$0.11 & 3.46$\pm$0.21 & 93.21$\pm$1.14 \\
    TURNIP (10-sec)  & 2.81$\pm$0.08 & 9.59$\pm$0.26 & 89.31$\pm$0.58 \\
    \hline
    \end{tabular}
\end{table}

In addition, we report our results using 10-sec time windows on the MMSE-HR dataset in Table~\ref{tab:10sec_mmsehr_results}, corresponding to the evaluation protocol we use on the MR-NIRP Car dataset. We believe that 10-sec windows are more appropriate for evaluation for several reasons: 1) Because 30-sec windows average the heart rate over a long duration, short-term errors in heart rate can be averaged out, making it seem as if a method estimates heart rate more accurately than it actually does. (In fact, some of the videos in the MMSE-HR dataset are not much longer than 30 sec.) 2) In many real-world applications, shorter wait times are more desirable or necessary. 3) The (more challenging) 10-sec window scenario more closely resembles a real-time, instantaneous measurement of heart rate, which may be important for clinical acceptance in the future. The results in Table~\ref{tab:10sec_mmsehr_results} show that for 10-sec windows, the MAE and RMSE are higher and the PTE6 is lower; this demonstrates that the 10-sec evaluation protocol is more challenging, with more room for performance gains that may lead to further improvements in algorithms.

Our algorithm outperforms previous methods on the near-infrared MR-NIRP Car dataset~\cite{autosparseppg} as well, as shown in Table \ref{tab:mrnirp_car_results}. Note that for this dataset, we use the OpenFace landmark detector~\cite{baltruvsaitis2016openface} rather than LUVLi~\cite{luvli}, and we average the landmark locations across 10 frames as described in~\cite{turnip}. On both the Driving (city driving) subset and the Garage (car running while parked in a garage) subset of the MR-NIRP Car dataset, we achieve significantly higher PTE6 than previous methods, indicating that our algorithm captures the true heart rate (within 6 bpm) a greater percentage of the time than previous methods. We also achieve smaller root-mean-squared error (RMSE) than previous methods on both subsets, showing that our method also reduces the error on a window-by-window basis. Even though the pulsatile signal is weaker in the NIR domain than in RGB, our method is still able to estimate the underlying pulse wave for heart rate estimation more accurately than previous methods.

\begin{table}[]
    \centering
    \caption{Results on the MR-NIRP Car dataset}
    \label{tab:mrnirp_car_results}
    \begin{tabular}{|c|cc|cc|}
    \hline
    \multirow{4}{*}{\textbf{Method}}& \multicolumn{4}{c|}{\textbf{MR-NIRP Car}} \\
    \cline{2-5}
    & \multicolumn{2}{c|}{Driving} & \multicolumn{2}{c|}{Garage} \\
     & \textbf{RMSE} & \textbf{PTE6}& \textbf{RMSE} & \textbf{PTE6} \\
     & (bpm) $\downarrow$ & (\%) $\uparrow$ & (bpm) $\downarrow$ & (\%) $\uparrow$ \\
    \hline
    
    DistancePPG \cite{Kumar2015DistancePPGRN} & $>$15 & 24.6 & $>$15 & 37.4  \\
    SparsePPG \cite{sparseppg} & $>$15 & 17.4 & $>$15 & 35.6 \\
    AutoSparsePPG \cite{autosparseppg} & 11.6 & 61.0 & 5.1 & 81.9 \\

    PhysNet-STSL-NIR \cite{inverseCAN} & 13.2 & 53.2 & 6.3 & 88.8 \\
   
    \hline
    \textbf{TURNIP}  & \textbf{11.4} & \textbf{65.1} & \textbf{4.6} & \textbf{89.7} \\
    \hline
    \end{tabular}
\end{table}

\subsubsection{Statistical Analysis of Heart Rate Estimation}
\begin{figure*}
\centering
\begin{subfigure}{.5\textwidth}
  \centering
  \includegraphics[width=.99\linewidth]{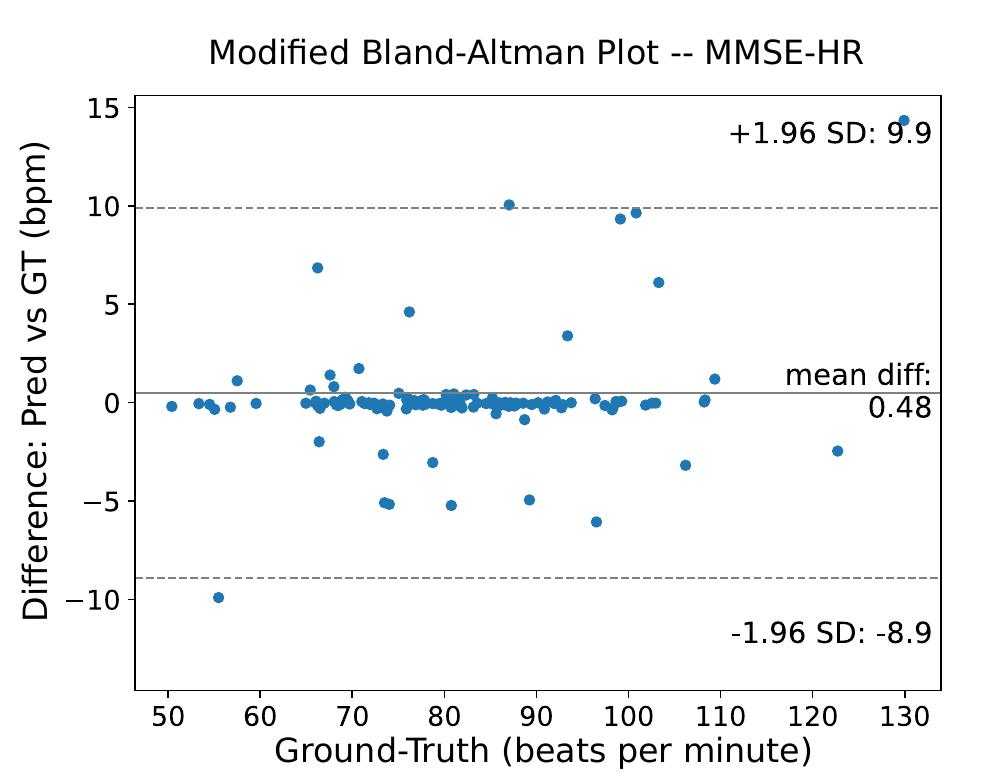}
  \caption{Bland-Altman analysis -- MMSE-HR Dataset}
  \label{fig:ba_mmse_hr}
\end{subfigure}
\begin{subfigure}{.5\textwidth}
  \centering
  \includegraphics[width=.99\linewidth]{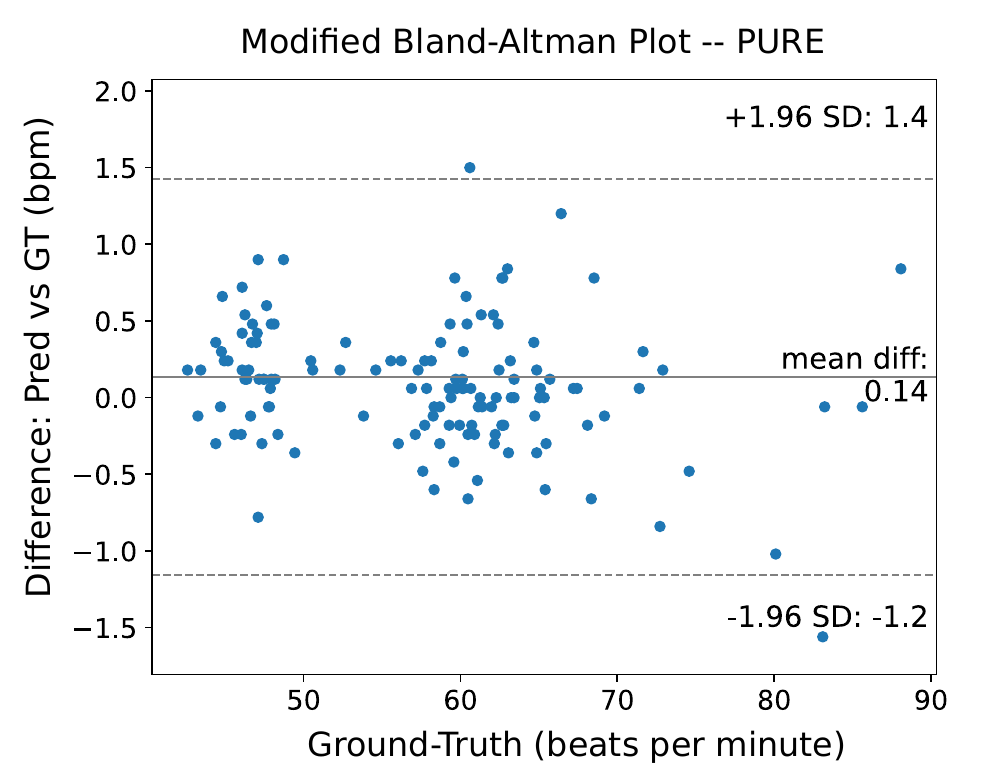}
  \caption{Bland-Altman analysis -- PURE Dataset}
  \label{fig:ba_pure_test10}
\end{subfigure}
\caption{Bland-Altman Analysis. Each point in the MMSE-HR and PURE graphs represent a non-overlapping 10-second window.}
\label{fig:ba_analysis}
\end{figure*}

\begin{figure*}[ht!]
     \centering
     \begin{subfigure}[b]{0.3\textwidth}
         \centering
         \includegraphics[width=\textwidth]{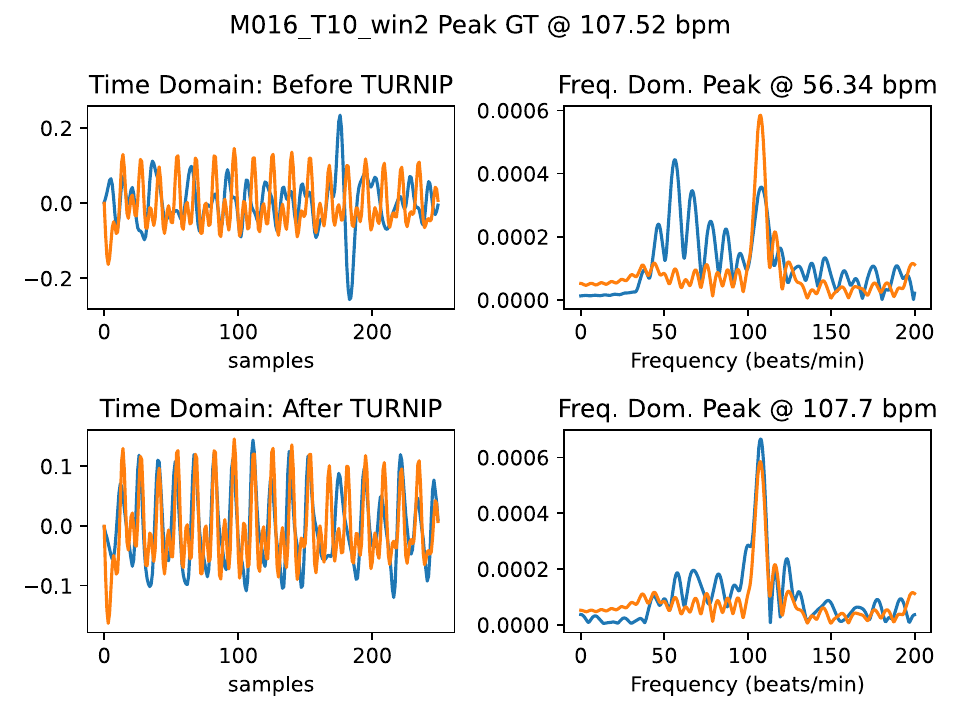}
         \caption{}
         \label{fig:M016_T10_win2_region2win2}
     \end{subfigure}
     \hfill
     \begin{subfigure}[b]{0.3\textwidth}
         \centering
         \includegraphics[width=\textwidth]{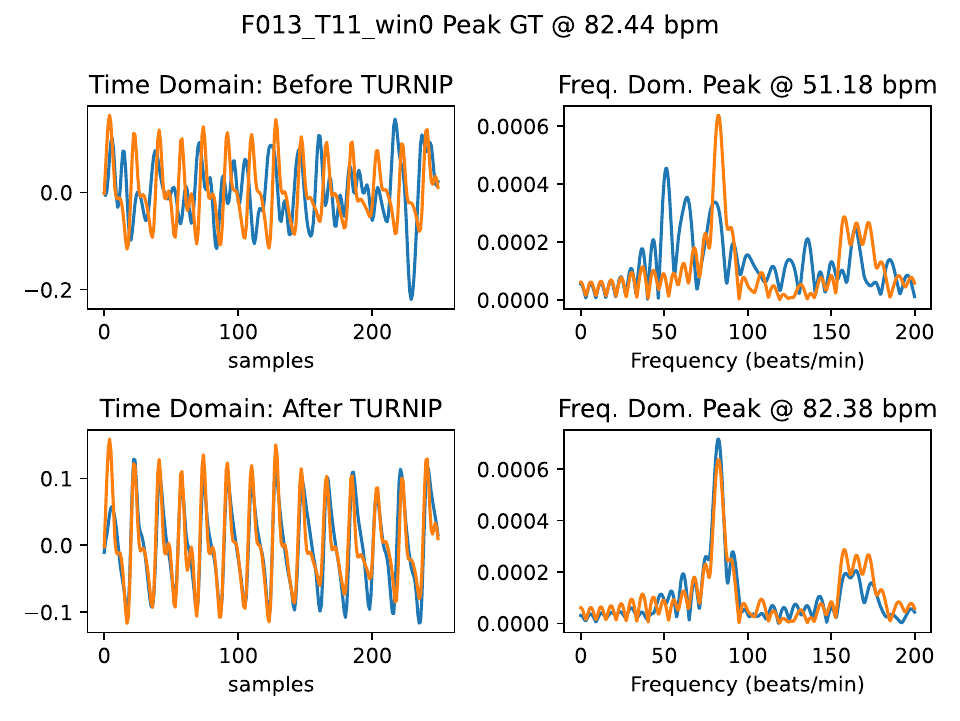}
         \caption{}
         \label{fig:F013_T11_win0_region2win2}
     \end{subfigure}
     \hfill
     \begin{subfigure}[b]{0.3\textwidth}
         \centering
         \includegraphics[width=\textwidth]{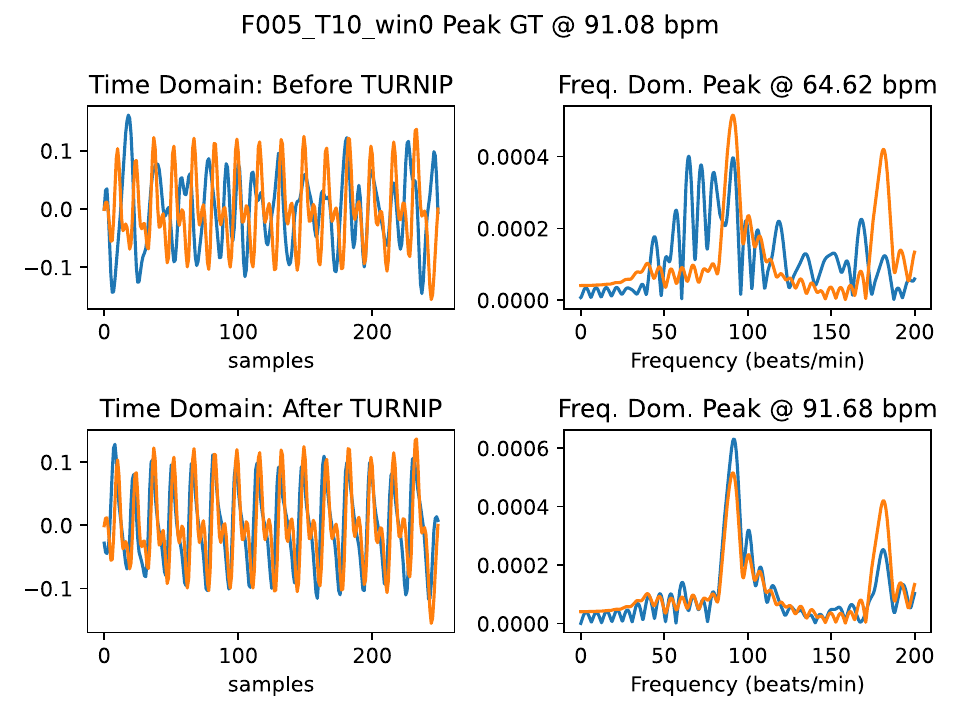}
         \caption{}
         \label{fig:F005_T10_win0_region0win0}
     \end{subfigure}
        \caption{iPPG estimation for three different signals The signals in orange are the ground-truth, with a peak frequency at 107.52 bpm in (a), 82.44 bpm in (b), and 91.08 bpm in (c), while the signals in blue are the inputs/outputs of our TURNIP pulse signal estimation algorithm.  The top and bottom rows respectively show the estimate \textit{before} and \textit{after} TURNIP pulse signal estimation. Each row shows the time-domain signals on the left and frequency-domain power spectra on the right. The title of each frequency spectrum plot  shows the peak frequency of the estimated signal.}
        \label{fig:three_iter}
\end{figure*}

We perform a modified Bland-Altman analysis --- plotting the ground-truth heart rate against the difference between the ground-truth and predicted heart rate --- for both the MMSE-HR dataset and PURE dataset in Figure~\ref{fig:ba_analysis}. On the MMSE-HR dataset, we see that we have a mean difference between predicted and ground-truth heart rates of 0.48, which shows that our predictions accurately match the ground-truth measurements. Furthermore, only two of our estimates fall outside of the 95\% limits of agreement, defined as 1.96 $\times$ the standard deviation of the differences. The greatest difference between the predicted and ground-truth heart rates occur at higher heart rate ranges, which most likely means that there was insufficient training data at those rates.

We do the same for the PURE dataset and show the results in Figure~\ref{fig:ba_analysis} on 10-second windows to show a variety of heart-rate estimates. In both cases, we see a mean difference very close to zero. Also, nearly all of our heart rate estimates fall within our limits of agreement, and the ones that are outside
our limits are still within 2 bpm of the ground truth.

\subsubsection{Pulse Rate Variability Analysis}
\begin{figure*}
    \centering
    \includegraphics[width=0.98\textwidth]{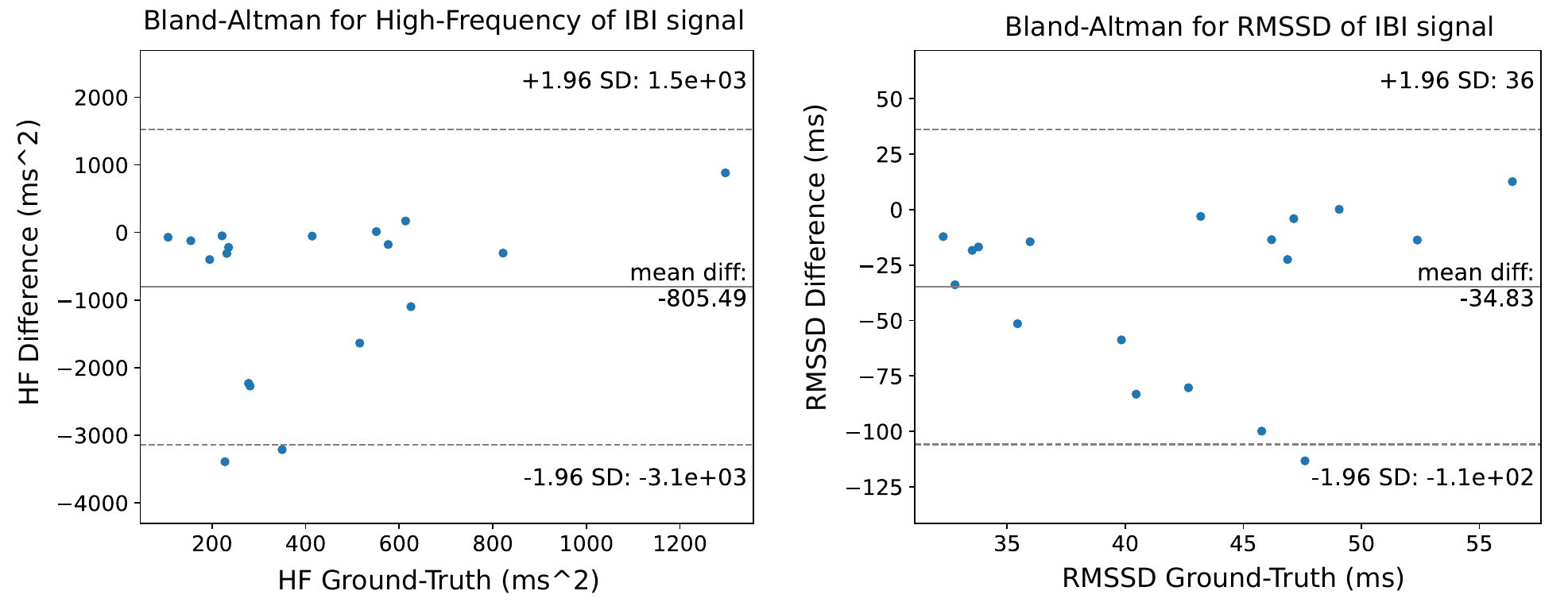}
    \caption{A Bland-Altman Analysis on the HF, and RMSSD metrics as described in Section~\ref{sec:hrv_analysis} on all 1-min videos of the PURE test set.} 
    \label{fig:hrv-ba}
\end{figure*}

We perform a HRV analysis on the predicted signals from TURNIP and the ground-truth and display the results in Figure~\ref{fig:hrv-ba}. We do this on the PURE dataset, for which videos are at at least 1-minute long. We note that our mean difference for the higher frequency component is -805.49 ms$^2$ while for the RMSSD metric our mean difference is -34.83 ms. For the high-frequency power estimate, many of our estimates have a difference close to zero, validating our ability to correctly predict the high frequency power. We notice a similar trend for the RMSSD metric. We believe that further research should focus on reconstructing waveform characteristics more effectively. 

\subsubsection{Qualitative Analysis}In Figure \ref{fig:three_iter}, each part $\big($(a), (b), or (c)$\big )$ shows example result waveforms for a single 10-sec time window from the test set of the MMSE-HR dataset. For each time window, the left column shows time-domain waveforms, and the right column shows the same signal in the frequency domain. In both the top and bottom rows, the signals in orange show the ground-truth pulse signal, while the overlaid signals in blue show the estimated signals. In the top row, the blue signal shows one channel (from one face region) of the output of our time series extraction module, \textit{before} TURNIP pulse signal estimation. In the bottom row, the blue signal shows our system's final estimate of the pulse signal, \textit{after} TURNIP pulse signal estimation. The peak frequency of each frequency domain graph provides the system's estimate of the heart rate.
 
From the frequency domain graphs (lower right of each part), we can see that our TURNIP pulse signal estimator closely reconstructs the underlying spectrum, attenuating spurious peaks and generating an accurate heart rate estimate. In Figure~\ref{fig:three_iter} (a), we see that the extracted time series signal (top row blue curves) has strong power at a lower frequency; our TURNIP algorithm attenuates this frequency and boosts the correct one. In (b), we see that the extracted time series signal is noisy in the frequency domain (blue curve at top right), with strong peaks at lower and higher frequencies than the true heart rate, and the predicted heart rate before TURNIP is greater than 6-beats away from the ground-truth. Our TURNIP pulse signal estimator attenuates the spurious peaks and correctly predicts the true heart rate (blue curve in bottom right). Figure~\ref{fig:three_iter} (c) shows similar behavior.

\begin{table}[]
    \centering
    \caption{Cross Dataset Comparison}
    \label{tab:cross_dataset}
    \begin{tabular}{|c|c|ccc|}
    \hline
    \multirow{2}{*}{\textbf{Train Dataset}} & \multirow{2}{*}{\textbf{Test Dataset}} & \textbf{MAE} & \textbf{RMSE} & \textbf{PTE6} \\
     & & (bpm) $\downarrow$ & (bpm) $\downarrow$ & (\%) $\uparrow$ \\
    \hline
    PURE & MMSE-HR & 4.35 & 13.54 & 89.14 \\
    MMSE-HR & PURE & 0.68 & 1.06 & 99.26\\
    \hline
    \end{tabular}
\end{table}

\subsection{Ablation studies}

\subsubsection{Cross-Dataset Evaluation}\label{sec:cross_datset} 
\begin{figure}
    \centering
    \includegraphics[width=0.47\textwidth]{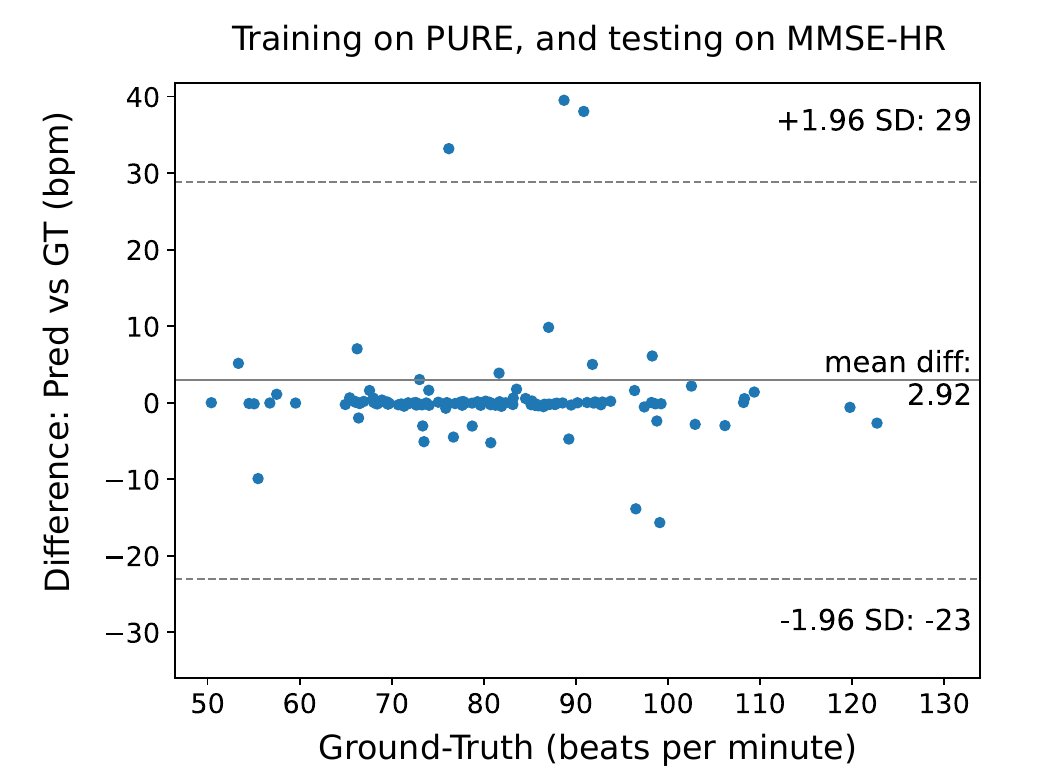}
    \caption{A modified Bland-Altman Analysis when training on PURE and testing on MMSE-HR. Each point represents a non-overlapping 10-second window of the MMSE-HR dataset. We see that we do well, except for a few large outliers.}
    \label{fig:pure_to_mmse_test}
\end{figure}

In this experiment, we train on one dataset and test on another dataset without any further fine-tuining; we show the results in Table~\ref{tab:cross_dataset}. We see that MMSE-HR transfers very well to the PURE dataset. We believe this occurs because of the nature of the MMSE-HR dataset -- firstly, there is more data upon which to train and secondly, it contains many examples of unconstrained motion with corresponding ground-truth signal from which the network can learn. The PURE dataset does not have the same distribution of unconstrained noise data. In addition to the significantly smaller dataset size, it does not contain the same noise characteristics that are typically found in the MMSE-HR dataset. Overall though, we still see good performance. When training on PURE and testing on MMSE-HR, we perform a Bland-Altman analysis and display the results in Figure~\ref{fig:pure_to_mmse_test}. We see that, on
average, we overestimate the ground-truth signal by approximately 3 beats per minute, but the majority
of our performance degradation results from a few outliers. Therefore, we see that the majority of
estimates, when training on PURE and testing on MMSE-HR, are within 5 beats of the true heart rate.

\subsubsection{Effect of Channel Ratio and Performance in low and high-motion conditions}
\begin{table}[]
        \centering
        \caption{Heart Rate estimation using different color channels and motion conditions}
        \label{tab:motion_results}
        \begin{tabular}{|c|c|ccc|}
        \hline
        \multirow{2}{*}{\textbf{Dataset}} & \multirow{2}{*}{\textbf{Motion-Channel}} & \textbf{MAE} & \textbf{RMSE} & \textbf{PTE6}  \\
         & & (bpm) $\downarrow$ & (bpm) $\downarrow$ & (\%)$\uparrow$ \\
        \hline
        \multirow{4}{*}{\textbf{MMSE-HR}} & Low Motion-Green   & 1.34  & 3.41 & 92.40  \\
        & Low Motion-RoG  &\textbf{0.80} & \textbf{2.33} & \textbf{96.20}  \\
        \cline{2-5}
        & High Motion-Green & 2.92 & 11.72 & 90.44 \\
        & High Motion-RoG & \textbf{1.92} & \textbf{5.02} & \textbf{92.50} \\
        \hline
        \multirow{4}{*}{\textbf{PURE}} & Low Motion-Green & 2.17 & 7.78 & 93.95  \\
        & Low Motion-RoG & \textbf{1.11} & \textbf{4.85} & \textbf{97.17}  \\
        
        \cline{2-5}
        & High Motion-Green & 2.23 & 7.27 & 94.33 \\
        & High Motion-RoG & \textbf{1.99} & \textbf{7.18} & \textbf{95.28} \\
        \hline
        \end{tabular}
    \end{table}

In Table~\ref{tab:motion_results}, we show the performance of TURNIP when the input channel is green or RedoverGreen (RoG) for low motion videos vs high motion videos. In the PURE dataset, we define as ``high motion" the videos labeled by the dataset authors as \textit{Slow Translation, Fast Translation,} and \textit{Medium Rotation}, while the ``low motion" videos are those labeled as \textit{Steady, Talking,} and \textit{Small Rotation}.

For each video in the MMSE-HR dataset, we measure the amount of motion in the video as follows. First, we compute the standard deviation across all frames of the 2D location of each of the 68 facial landmarks found by LUVLi~\cite{luvli}. Next, we compute the mean of these 68 standard deviations to obtain a single scalar measure of motion in the video. For each subject, the one video with the most motion is considered ``high motion," and the remaining videos of that subject are considered to be ``low motion." See the appendix for the exact splits. As we can see from the table, in all scenarios the RoG channel improves heart-rate estimation performance, whether it be high motion or low motion. This shows the effectiveness of our use of color channels.

\begin{table}[!h]
    \centering
    \caption{Color Channel performance on the MMSE-HR dataset}
    \label{tab:multiplechannel}
    \begin{tabular}{|c|ccc|}
    \hline
    \multirow{2}{*}{\textbf{Method}} & \textbf{MAE} & \textbf{RMSE} & \textbf{PTE6} \\
     & (bpm) $\downarrow$ & (bpm) $\downarrow$ & (\%) $\uparrow$ \\
    \hline
    Red & 3.66$\pm$0.87 & 9.22$\pm$1.32 & 82.94$\pm$0.44 \\
    Blue & 44.93$\pm$2.31 & 47.33$\pm$4.01 & 1.55$\pm$0.52 \\
    Green  & 1.63$\pm$0.21 & 5.93$\pm$1.11 & 90.88$\pm$0.84\\
    \{Red, Green\} & 1.81$\pm$0.31 & 5.47$\pm$1.31 & 87.72$\pm$0.48 \\ 
    \{Green, Blue\} & 2.45$\pm$0.61 & 8.03$\pm$1.05 & 89.14$\pm$0.33 \\ 
    \{Red, Blue\} & 3.04$\pm$0.55 & 8.45$\pm$1.75 & 85.27$\pm$0.61 \\
    Red-over-Green (R/G) & \textbf{1.17$\pm$0.11} & \textbf{3.46$\pm$0.21} & \textbf{93.21$\pm$1.14}\\
    \{Red, Green, Blue\} & 2.93 $\pm$0.18 & 10.13$\pm$0.47 & 89.14 $\pm$0.15\\
    \hline
    \end{tabular}
\end{table}

\subsubsection{Using multiple color channels} In Table~\ref{tab:multiplechannel}, we experiment with different color channels as input to TURNIP, stacking the color channels into a matrix of size $T \times 48 \times 3$ for  \{Red,Green,Blue\} input or $T \times 48 \times 2$ for 2-color input such as \{Red,Green\}. When we include the other color channels, including the weaker blue signal, we introduce in-domain noise that makes optimization more difficult. The 3-channel R,G,B TURNIP can not replicate the more-optimal ratio of the red and green channels, nor does it learn how to handle the noise as effectively. We notice the best results when inputting the red over green channel directly, and report these results. This shows that instead of using deep learning-based Spatio-Temporal maps, a simple modification to the signal extraction procedure in the RGB image domain can result in state-of-the-art performance.

\begin{table}[]
    \centering
    \caption{Understanding effect of occlusion handling}
    \label{tab:occlu}
    \begin{tabular}{|c|ccc|}
    \hline
    \multirow{2}{*}{\textbf{Occlusion Handling?}} & \textbf{MAE} & \textbf{RMSE} & \textbf{PTE6} \\
     & (bpm) $\downarrow$ & (bpm) $\downarrow$ & (\%) $\uparrow$ \\
    \hline
    \xmark  & 1.21$\pm$0.07 & 3.53$\pm$0.24 & 93.01$\pm$0.95 \\
    \cmark & \textbf{1.17$\pm$0.11} & \textbf{3.46$\pm$0.21} & \textbf{93.21$\pm$1.14} \\
    \hline
    \end{tabular}
\end{table}

\subsubsection{Handling self-occluded landmarks}One of our contributions involves the detection and handling of self-occluded and out-of-frame landmarks during training, which improves model robustness during inference. We show in Table \ref{tab:occlu} the effects of incorporating or omitting the occlusion handling module. On the challenging MMSE-HR dataset, using our occlusion handling improves performance, increasing PTE6 from 93.02 to 93.21. 
\begin{table}[]
    \centering
    \caption{Effects of data augmentation and GRU on TURNIP performance using the MR-NIRP Car dataset (10-second windows) and MMSE-HR dataset (30-second windows).}
    \label{tab:mrnirp_car_ablation}
    \begin{tabular}{|cc|cccc|cc|}
    \hline
    \multicolumn{2}{|c|}{\multirow{2}{*}{\textbf{Method}}}& \multicolumn{4}{c|}{\textbf{MR-NIRP Car}} & \multicolumn{2}{c|}{\textbf{MMSE-HR}} \\
    & & \multicolumn{2}{c}{Driving} & \multicolumn{2}{c|}{Garage} &  &\\
    Aug. & GRU & RMSE & PTE6 & RMSE & PTE6 & RMSE & PTE6\\
    \hline
    
    \xmark & \cmark & \textbf{10.7} & 61.9 & 5.9 & 81.9 & 5.72 & 91.47\\
    \cmark & \xmark & 11.4 & 63.3 & 5.0 & 89.7 & 3.73 & 89.92\\
    \hline
    \cmark & \cmark & 11.4 & \textbf{65.1} & \textbf{4.6} & \textbf{89.7} & \textbf{3.46} & \textbf{93.21}\\
    
    \hline
    \end{tabular}
\end{table}

\subsubsection{Effects of Gated Recurrent Unit and Data Augmentation}Table \ref{tab:mrnirp_car_ablation} shows the effect of including or omitting the GRU component in our spatio-temporal denoising U-Net. We clearly see that the GRU plays an important role in improving the performance.
The table shows that our data augmentation generally improves results on the MR-NIRP Car dataset. We see that data augmentation improves the PTE6 on the MMSE-HR dataset and both subsets of the MR-NIRP Car dataset,  and it improves RMSE on all but the the ``Driving" subset of MR-NIRP Car. The data augmentation and GRU both help our method outperform the previous methods on the MR-NIRP Car dataset, as shown in Table~\ref{tab:mrnirp_car_results}.

\begin{table}[]
        \centering
        \caption{Evaluating ICA/CHROM/POS using our signal extraction pipeline on the MMSE-HR dataset}
        \label{tab:re-implementation}
        \begin{tabular}{|c|ccc|}
        \hline
        \multirow{2}{*}{\textbf{Method}} & \textbf{MAE} & \textbf{RMSE}  & \textbf{PTE6} \\
         & (bpm) $\downarrow$ & (bpm) $\downarrow$ & (bpm) $\uparrow$ \\
         \hline
         ICA \cite{Poh_2010} & 5.44 & 12.00 & - \\
         ICA (our pipeline) & 7.62 & 16.57 & 74.41 \\
         \hline
        CHROM \cite{chrom} & 3.74 & 8.11 & -\\
        CHROM (our pipeline) & 2.84 & 9.77 & 88.97\\
        \hline
        POS \cite{algorithmic_principles} & 3.90 & 9.61 & - \\
        POS (our pipeline) & 4.08 & 11.25 & 82.94 \\
        
        \hline
        \hline
        
        TURNIP & \textbf{1.17} & \textbf{3.46}  & 93.21 \\
        \hline
        \end{tabular}
    \end{table}

\subsubsection{Effects of our Signal Extraction on ICA/POS/CHROM} We evaluate the performance of ICA/POS/CHROM on our custom signal extraction pipeline (Face Detection + LUVLi Landmark Detection + Time-Series Extraction) and display the results in Table~\ref{tab:re-implementation}. We note a few differences in
implementation: as compared to TURNIP which used 48 subregions (which were later collapsed into one through convolutional layers), the original ICA/CHROM/POS used the entire face as a single region for
estimation. For our implementation of ICA/CHROM/POS, we extracted 48 time-series from the video; to capture the heart rate, we converted each region into it’s power spectrum, summed each frequency
bin across all 48 regions, and selected the bin with the highest power as our heart rate estimate. We note that TURNIP has superior performance under the same signal extraction technique, showing the validity of our method.

\section{Conclusion}\label{sec:conclusion}
In this paper, we build a modular pipeline for non-contact heart rate estimation and pulse rate variability from facial videos that is composed of modules that perform face and landmark detection, time series extraction, and pulse signal estimation. Compared to previous end-to-end deep-network methods that map directly from the RGB frames to the final output, our modular algorithm significantly improves the estimation results. Our novel handling of self-occluded and out-of-frame landmarks in our time series extractor and TURNIP pulse signal estimator make our algorithm robust to varying levels of occlusion. Additionally, our signal model uses the ratio of pixel intensities of the red channel to the green channel for RGB videos, and leads to significant improvements across metrics.  Our results demonstrate a new state-of-the-art in estimating pulse signals from facial videos, and our model achieves this using stages that are modular and interpretable. We have tested our algorithm in two different imaging domains (RGB and near-infrared), outperforming previous methods in both domains.

Future work should continue to address real-world remote heart rate and pulse rate variability estimation with increasing variation in illumination and motion. While this work and many previous works address some of the fundamental issues associated with illumination variation and motion, further research could help make algorithms even more robust to these issues.

\section*{Appendix}

    In Table VII, we tested our algorithm on high-motion and low-motion splits of the PURE and MMSE-HR datasets. We enumerate the data splits below.

    \begin{itemize}
    \item \textbf{High-motion (PURE)}: Videos labeled as {03-Slow Translation, 04-Fast Translation, and 06-Medium Rotation}.
    \item \textbf{Low-Motion (PURE)}:Videos labeled as {01-Steady, 02-Talking, and 05-Small Rotation}
\end{itemize}

    \begin{itemize}
    \item \textbf{High-Motion (MMSE-HR)}: F005-T10, F006-T11, F007-T11, F008-T10, F009-T11, F010-T11, F011-T11, F012-T11, F013-T8, F014-T8, F015-T8, F016-T8, F017-T8, F018-T1, F019-T11, F020-T10, F021-T11, F022-T10, F023-T10, F024-T10, F025-T10, F026-T10, F027-T10, M001-T11, M002-T11, M003-T10, M004-T10, M005-T10, M006-T10, M007-T10, M008-T11, M009-T10, M010-T11, M011-T8, M012-T11, M013-T11, M014-T10, M015-T10, M016-T11, M017-T10
    \item \textbf{Low-motion (MMSE-HR)}: F005-T11, F006-T10, F007-T10, F008-T11, F009-T10, F010-T10, F011-T10, F012-T10, F013-T10, F013-T11, F013-T1, F014-T10, F014-T11, F014-T1, F015-T10, F015-T11, F015-T1, F016-T10, F016-T11, F016-T1, F017-T10, F017-T11, F017-T1, F018-T10, F018-T11, F018-T8, F019-T10, F020-T11, F020-T1, F020-T8, F021-T10, F022-T11, F022-T1, F022-T8, F023-T11, F024-T111, F025-T11, F026-T11, F027-T11, M001-T10, M002-T10, M003-T11, M004-T11, M005-T11, M006-T10, M007-T11, M008-T10, M009-T11, M010-T10, M011-T11, M012-T10, M013-T10, M014-T11, M016-T11, M016-T10, M017-T11
\end{itemize}

\bibliographystyle{IEEEtran}
\bibliography{ref}

\end{document}